\documentclass{article}
\usepackage{spconf,amsmath,graphicx}

\usepackage{graphicx}
\usepackage{booktabs}
\usepackage{flexisym}
\usepackage{multicol}
\usepackage{gensymb}
\usepackage{verbatim}
\usepackage{cite}
\usepackage{amssymb}
\usepackage{xcolor}
\usepackage{multirow}
\usepackage{bm}
\usepackage{setspace} 
\usepackage{csquotes}

\usepackage{enumitem}
\setlist{nosep, leftmargin=14pt}

\usepackage{mwe} 

\title{Self-supervised Graph Transformer with Contrastive Learning for Brain Connectivity Analysis towards Improving Autism Detection}
\name{Yicheng Leng$^{1}$, Syed Muhammad Anwar$^{2}$, Islem Rekik$^{3}$, Sen He$^{4}$, Eung-Joo Lee$^{1,*}$}
\address{$^{1}$Department of Electrical and Computer Engineering, University of Arizona, Tucson, USA\\
$^{2}$Children's National Hospital, Washington, DC, USA\\
$^{3}$Department of Computing, Imperial College London, London, UK\\
$^{4}$Department of Systems and Industrial Engineering, University of Arizona, Tucson, USA\\
}
\begin{document}


%
\maketitle
\begin{abstract}
Functional Magnetic Resonance Imaging (fMRI) provides useful insights into the brain function both during task or rest. Representing fMRI data using correlation matrices is found to be a reliable method of analyzing the inherent connectivity of the brain in the resting and active states. Graph Neural Networks (GNNs) have been widely used for brain network analysis due to their inherent explainability capability. In this work, we introduce a novel framework using contrastive self-supervised learning graph transformers, incorporating a brain network transformer encoder with random graph alterations. The proposed network leverages both contrastive learning and graph alterations to effectively train the graph transformer for autism detection. Our approach, tested on Autism Brain Imaging Data Exchange (ABIDE) data, demonstrates superior autism detection, achieving an AUROC of $82.6$ and an accuracy of 74\%, surpassing current state-of-the-art methods.

\end{abstract}

\section{Introduction}
Autism spectrum disorder (ASD) is a complex neurodevelopmental condition characterized by challenges in social interaction, communication, and restricted or repetitive behavior. The clinical heterogeneity of ASD and its overlap with other neurological disorders underscore the need for objective and reliable diagnostic methods~\cite{dawson2019potential}. Traditional diagnostic approaches are largely based on behavioral and neuro-psychological assessments, which are subjective and dependent upon the clinician's experience and the patient's condition during the assessment.

Neuroscientists have actively studied brain network analysis to comprehend human brain organization, treat disorders, and predict clinical outcomes~\cite{deco2011emerging}. Recent brain imaging studies highlight the importance of interactions between brain regions for neural development and disorder analysis~\cite{li2021braingnn}. Among various imaging techniques, functional Magnetic Resonance Imaging (fMRI) is widely used to assess brain function and connectivity networks. Functional connectivity captures correlations in time-series signals between brain regions, while structural connectivity represents physical links among the cortical and subcortical areas~\cite{maglanoc2020multimodal}. Using graph theory, brain networks represented with nodes and edges can be formulated to depict brain region interactions~\cite{farahani2019application}. In particular, nodes in the network are defined as Regions of Interest (ROIs) based on the brain atlas used. Edges are calculated as pairwise correlations between the Blood-Oxygen-Level-Dependent (BOLD) signal series obtained from each region~\cite{smith2011network}. This network pattern facilitates the classification of brain regions into diverse functional modules, thereby aiding in disease diagnosis, studying brain development, and enhancing understanding of disease progression.

Graph neural networks and connectivity analysis provides great potential to study and understand complex brain disorders such as autism~\cite{anirudh2019bootstrapping}. Efficient training of such networks is challenging, as high-quality labels are required for most deep learning models. Self-Supervised Learning (SSL) is an alternative paradigm, where models can be trained with a pre-text task such as image reconstruction, alleviating the burden of collecting high-quality labels. Once trained, such models can be used for downstream clinical tasks with few labeled data~\cite{anwar2023spcxr}. We build on the strengths of both SSL and graph neural networks and propose a novel training paradigm for autism detection using fMRI data. Particularly, we focus on contrastive learning, with a novel graph dilation/shrinkage strategy to detect autism. 

\noindent{\textbf{Our contributions:}} The main contributions of our proposed methods are as follows:
\begin{itemize}
    \item We develop a self-supervised learning strategy for graph transformer utilizing brain network transformer encoder.
    \item We introduce a graph dilation and shrinkage strategy to improve contrastive learning and achieve better detection outcomes.
    \item We present extensive experiments, demonstrating state-of-the-art detection performance (AUROC of 82.6), to showcase the effectiveness of our proposed method using contrastive self-supervised learning with noise addition.
\end{itemize}
\color{black}

\begin{figure*}[h]
\begin{center}
    \includegraphics[width=\linewidth]{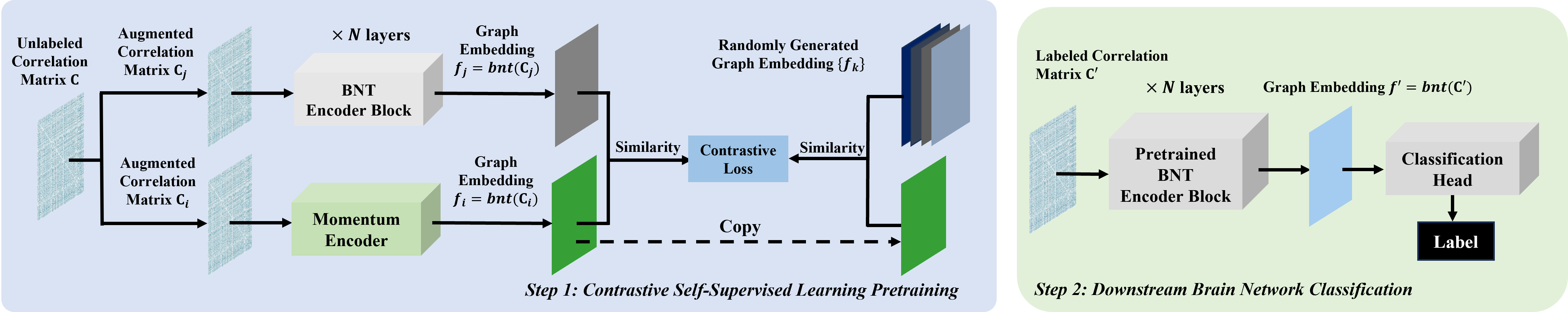}
\end{center}
\vspace{-3mm}
   \caption{Overview of our proposed contrastive self-supervised learning framework.
   }
\label{fig_method1}
\end{figure*}
\vspace{-3mm}

\section{Related Work}
\noindent{\textbf{Autism Detection and Analysis}}
Functional magnetic resonance imaging, even in the resting state (rs-fMRI) and the associated Functional Connectivity (FC) matrices have opened new avenues for understanding neurological development and disorders, offering a robust and accurate tool to identify disease biomarkers~\cite{lee2013resting}. FC captures the synchronicity of behavioral patterns across different regions of the brain at rest, revealing the functional architecture. Typically, FC in patients shows a reduction in long-range communication between the frontal and posterior brain regions, accompanied by an increase in local connections. These changes can lead to disrupted cortical connectivity in autism~\cite{nomi2015developmental}. However, FC studies on ASD reveal variable changes in FC, with both increased and decreased connectivity, leading to inconsistent findings. Deep learning have proven to be highly effective in detecting subtle patterns in high-dimensional data that are often imperceptible to human observation~\cite{dawson2019potential}. 

\noindent{\textbf{GNNs for Brain Connectivity Analysis}} \label{works}
Recent research has focused on extending Graph Neural Network (GNN) models to analyze brain connectivity. For instance, brainGNN leverages ROI-aware GNNs to utilize functional information and employs special pooling operators for node selection~\cite{li2021braingnn}. FBNetGen explores the generation of brain networks and their explainability for downstream tasks~\cite{kan2022fbnetgen}. Additionally, Graph Transformers have shown strong performance in graph representation learning, integrating edge information and using eigenvectors for positional embeddings ~\cite{ying2021transformers}. The Spectral Attention Network (SAN) improves positional embeddings and refines the attention mechanism by prioritizing neighboring nodes alongside global information~\cite{kreuzer2021rethinking}. The Nrain Network Transformer (BNT) models brain networks as graphs with fixed-size and ordered nodes and introduces a graph Transformer model with an orthonormal clustering readout for brain network analysis~\cite{kan2022brain}. 

GNN training is challenging and unstable with limited datasets, leading to high variance and poor performance. Also, more complex GNN models tend to cause greater variance and lower performance~\cite{zhu2022joint}. Our strategy mitigates this problem, offering stable learning outcomes with lower variance and state-of-the-art performance for autism detection.

\section{Methods}
Our proposed framework is illustrated in Figure~\ref{fig_method1}. We introduce an innovative method for pretraining the BNT encoder~\cite{kan2022brain} using contrastive loss to enhance feature learning. Subsequently, we use the pretrained BNT encoder for supervised training in downstream brain network classification for autism detection.
\vspace{-3mm}
\subsection{Contrastive Self-supervised Learning (CSSL) with Brain Network Transformer}
In the fMRI datasets, each brain network comprises time series data represented as $\mathbf{T}\in\mathbb{R}^{V\times1}$ and a connectivity profile denoted by the correlation matrix $\mathbf{C}\in\mathbb{R}^{V\times V}$, where $V$ indicates the number of nodes (i.e., ROIs). In particular, the BNT utilizes $\mathbf{C}$ as inputs for brain network classification, employing orthonormal clustering readout functions to effectively discern modular-level similarities among ROIs in brain networks.

In this work, we propose a framework based on Contrastive Self-Supervised Learning (CSSL)~\cite{zeng2021contrastive} to pretrain the BNT encoder $bnt(\cdot)$, as illustrated in Fig.~\ref{fig_method1}. The key idea involves using the CSSL framework to generate new instances from unlabeled networks and employing the BNT encoder to identify whether two generated networks are from the same original instance. Using CSSL, we improve the model to facilitate effective learning of brain network representations, thus enhancing classification performance during the downstream task. In addition, our framework incorporates graph dilation and shrinkage techniques, specifically designed for brain network analyses. The proposed method allows for generating new instances and improving representation learning, as detailed in the following section.
\vspace{-3mm}
\subsection{Graph Dilation and Shrinkage}
Basic graph augmentation includes node and edge additions and deletions. When generating new instances for contrastive learning through these random operations, the BNT encoder learns self-attention based on the augmented graph structure. However, in brain networks, ROIs are fully connected, with connectivity represented by the correlation matrix $\mathbf C$. Thus, we revise the graph augmentation to focus solely on edge operations since adding or deleting nodes is not applicable in our use case. Intuitively, the correlation between two nodes indicates their connectivity. Hence, we modify edge addition and deletion to increase or decrease the absolute value of correlation, with the increment being random. Furthermore, to simulate node operations, we randomly increase or decrease all adjacent edges of randomly selected nodes, altering connectivity within the ROI. This is termed \textit{graph dilation and shrinkage}.

\vspace{-3mm}
\begin{figure}[h]
\begin{center}
    \includegraphics[width=0.95\linewidth]{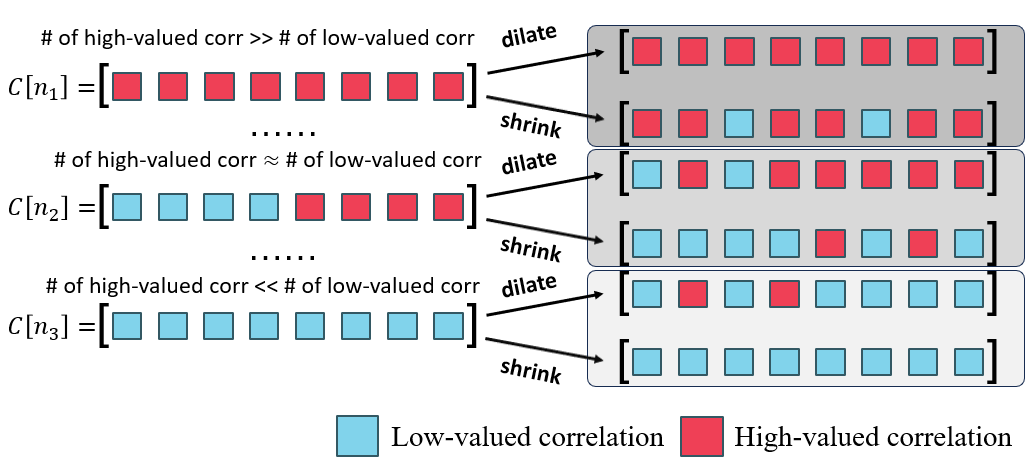}
\end{center}
\vspace{-3mm}
   \caption{Graph dilation and shrinkage on correlation matrix $\mathbf{C}$ with nodes $n_i$.
   }
\label{fig_method2}
\end{figure}
\vspace{-3mm}

Figure~\ref{fig_method2} illustrates three examples of graph dilation and shrinkage proposed in this work. For nodes $n_1$ and $n_2$, if the \enquote{shrink} operation results in new vectors with low-valued correlations, it resembles the process of deleting a node. For node $n_3$, if the \enquote{dilate} operation results in new vectors comprising high-valued correlations, it simulates the process of node addition. In other cases, the dilation and shrinkage are similar to edge addition and deletion, respectively. Note that the \enquote{dilate} operation for $n_1$ and \enquote{shrink} operation for $n_3$ are not without importance because the connectivity values, which describe the correlation, are not binary. This potentially contains information for the encoder to learn. Additionally, to avoid over-fitting, we add a Gaussian noise with $N(0,0.01)$ to the nodes not selected for dilation and shrinkage. In summary, our graph dilation and shrinkage simulate all graph augmentation operations in brain networks, suitable for constructing the CSSL framework.

The predictive task of CSSL is to train the BNT encoder to predict whether two embeddings of graphs, $f_1 = bnt(C_1)$ and $f_2 = bnt(C_2)$, are from the same graph. This involves generating two augmented graphs, $C_i$ and $C_j$, along with a set of randomly generated graphs $\{C_k\}$ for each unlabeled graph $C$. With that, and the training set $X = \{ x_1, \ldots, x_N \}$, the contrastive loss is formulated as follows:
\begin{equation}
    L(x_i)=-log \frac{e^{(sim(g(f_i), g(f_j))/\tau)}}{e^{(sim(g(f_i), g(f_j))/\tau)} + \sum_k e^{(sim(g(f_i), g(f_k))/\tau)}},
    \label{eq_method1}
\end{equation}
where $sim$ represents the cosine function. Similar to the approach used in CSSL~\cite{zeng2021contrastive}, we employ MoCo~\cite{he2020momentum} to optimize Equation~\ref{eq_method1}. Using CSSL, we load the pretrained parameters for the BNT encoder to conduct finetuning on the BNT model, incorporating a new classification head identical to the one used in BNT. 

\section{Experiments}
\subsection{Experimental Settings}
\subsubsection{Datasets} There are a limited number of brain imaging datasets that contain raw imaging data for preprocessing to generate brain network datasets, primarily due to regulatory restrictions. To address this, we use the preprocessed Autism Brain Imaging Data Exchange (ABIDE) dataset, an openly accessible compilation of rs-fMRI data from autistic spectrum disorder subjects and neurotypical controls~\cite{craddock2013neuro}. The dataset spans a broad age range and includes data from numerous healthy and ASD participants, collected over an extended period of time. In specific, the pre-processed data consist of 1,009 samples, each comprising 200 nodes and 40,000 edges. The node features consist of time series data with a length of 100, resulting in a 200$\times$100 matrix for every sample. Moreover, the edge features, represented by the correlations, result in a 40,000$\times$1 matrix for each sample. With that, the correlation matrix is dense, and these characteristics remain consistent across all samples within the ABIDE dataset.
\begin{figure*}[t]
\begin{center}
    \includegraphics[width=0.9\linewidth]{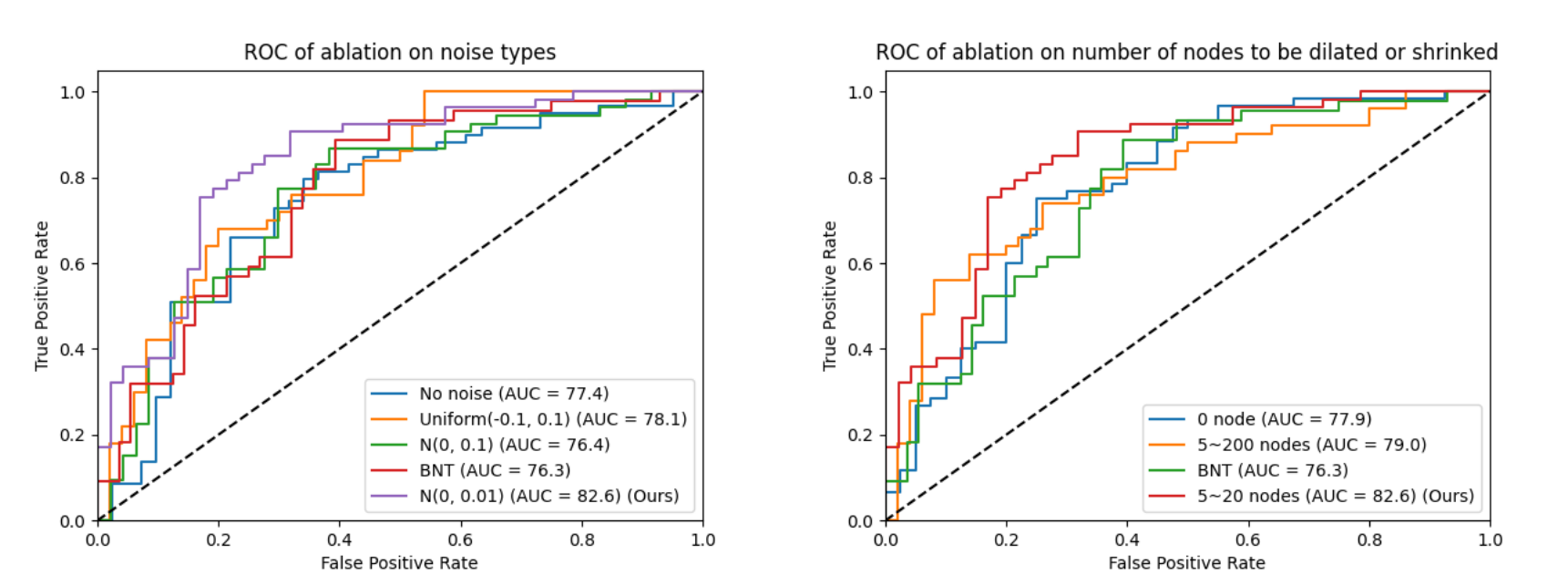}
\end{center}
\vspace{-3mm}
   \caption{ROC curves for various experiments on the variation of number of nodes selected for graph dilation/shrinkage as well as noise types and levels. The red curve shows our selected settings with state-of-the-art performance. 
   }
\label{fig_ROC}
\end{figure*}
\vspace{-3mm}

\subsubsection{Evaluation Metrics}
For this binary classification task, we used AUROC and accuracy (threshold set at 0.5) as evaluation metrics, along with sensitivity and specificity to assess model performance. Specificity denotes the ratio of accurately classified negative samples, whereas sensitivity represents the ratio of correctly classified positive samples.

\vspace{-3mm}
\subsubsection{Implementation Details}
For the MoCo implementation, we follow the CSSL approach~\cite{zeng2021contrastive}, setting the queue size to 512, momentum to 0.999, and temperature $\tau$ to 0.07. We use the stochastic gradient descent optimizer with a learning rate of 0.00001 and a batch size of 64. We perform CSSL pretraining for 900 epochs for our method and ablations, using all instances in ABIDE without labels. For finetuning, we combine the pretrained BNT encoder with a classification head for the downstream graph classification, as in BNT~\cite{kan2022brain}. At the end of the BNT encoder, the features are reshaped to dimensions ($N_b$, $N_o$, 8), where $N_o$ is the output number of nodes = 100. The classification head includes Linear($D_f$, 256), LeakyReLU, Linear(256, 32), LeakyReLU, Linear(32, 2), with $D_f$=8$\times$$N_o$, with $N_b$ and $D_f$ representing the batch size and the flattened feature dimension.

\begin{table}[h]
  \centering
  \caption{Results of performance comparison using various GNN based methods.}
  \resizebox{0.47\textwidth}{!}{\setlength{\tabcolsep}{1mm}
  {
    \begin{tabular}{ccccc}
    \toprule
    Method & Accuracy & AUROC & Sensitivity & Specificity \\
    \midrule
    BrainGNN~\cite{li2021braingnn} & 59.4±2.3 & 62.4±3.5 & 36.7±24.0 & 70.7±19.3 \\
    BrainGB~\cite{cui2022braingb} & 63.6±1.9 & 69.7±3.3 & 63.7±8.3 & 60.4±10.1 \\
    FBNetGen~\cite{kan2022fbnetgen} & 68.0±1.4 & 75.6±1.2 & 64.7±8.7 & 62.4±9.2 \\
    Graphormer~\cite{ying2021transformers} & 60.8±2.7 & 63.5±3.7 & \textbf{78.7±22.3} & 36.7±23.5 \\
    SAN~\cite{kreuzer2021rethinking}   & 65.3±2.9 & 71.3±2.1 & 55.4±9.2 & 68.3±7.5 \\
    BNT~\cite{kan2022brain}   & 69.6± 3.8 & 76.3±3.0 & 72.5±11.1 & 67.3±12.7 \\
    \midrule
    \textbf{Ours}  & \textbf{74.4±2.4} & \textbf{82.6±1.8} & 66.9±8.3 & \textbf{81.7±3.6} \\
    \bottomrule
    \end{tabular}%
    }}
    
  \label{tab_comparison}%
\end{table}%

We train the BNT model for 200 epochs using the Adam optimizer, with a learning rate of 0.00005, weight decay of 0.00005, and a batch size of 64. We use a relatively smaller learning rate than BNT to ensure stability of performance. Similar to the approach used in ~\cite{kan2022brain}, we use a stratified sampling strategy during the training-validation-testing splitting process to achieve more stable performance, with 70\% for training, 10\% for validation and the remainder for testing. For each experimental result, we repeat the above process five times and record the mean and standard deviation of the results. In each iteration, we select the model with the highest validation AUROC to record the test metrics for comparison.

\subsection{Results}
\subsubsection{Performance Analysis}
Table~\ref{tab_comparison} presents the results of our model. We compare our results with existing GNN-based methods, including the Graph Transformer from Section~\ref{works}. In general, our proposed method achieved an AUROC of 82.6 $\pm$ 1.8, which is the highest compared to other methods in the literature. The standard deviation also shows that the models results are stable, which is important for reliable test time performance. In this work, we reproduce the state-of-the-art results achieved by the BNT method and compare our performance with those reported by them~\cite{kan2022brain}.
\vspace{-3mm}
\begin{table}[h]
  \centering
  \caption{Ablation studies on the number of nodes for graph augmentation and different types and levels of noise.}
    \resizebox{0.47\textwidth}{!}{\setlength{\tabcolsep}{1mm}\begin{tabular}{c|c|cccc}
    \toprule
    \multicolumn{2}{c|}{Ablative settings} & Accuracy & AUROC & Sensitivity & Specificity \\
    \midrule
    The number  &0     & 68.6$\pm$4.5 & 77.9$\pm$2.3 & 65.9$\pm$15.5 & 70.5$\pm$7.9 \\
    of nodes&5$\sim$200 & 71.2$\pm$2.5 & 79.0$\pm$3.0 & \textbf{66.9$\bm{\pm}$7.9} & 74.6$\pm$8.2 \\
    \midrule
     &Without Noise &  68.0$\pm$ 5.5     &   77.4$\pm$5.8   &  72.9$\pm$14.4     & 62.8$\pm$19.5 \\
    Noise&Uniform(-0.1, 0.1) & 71.4$\pm$ 2.3 & 78.1$\pm$3.6 & \textbf{75.3$\bm{\pm}$8.7} & 67.8$\pm$4.8 \\
    &N(0, 0.1) & 70.6$\pm$ 5.1 & 76.4$\pm$3.2 & 71.7$\pm$13.1 & 70.1$\pm$5.8 \\
    \midrule
    \multicolumn{2}{c}{\textbf{Ours (5$\sim$20, N(0, 0.01))}} & \textbf{74.4$\bm{\pm}$2.4} & \textbf{82.6$\bm{\pm}$1.8} & 66.9$\pm$8.3 & \textbf{81.7$\bm{\pm}$3.6} \\
    \bottomrule
    \end{tabular}}%
    
  \label{tab_ablation_number}%
\end{table}%
\vspace{-5mm}

\subsubsection{Ablation Studies}
This section presents results from various configurations of our training strategy. The ablation study is conducted to explore the impact of different numbers of nodes for graph augmentation and different types and levels of noise applied during training. The top-performing results in each metric are highlighted in bold. Table~\ref{tab_ablation_number} presents performance comparisons using different numbers of nodes for graph dilation and shrinkage, showing that excessive node usage can lead to overfitting and reduced performance. Table~\ref{tab_ablation_number} also provides comparisons using different types of noise. The corresponding ROC curves are presented in \ref{fig_ROC}. The results indicate that low-variance Gaussian noise improves classification accuracy, AUROC, and specificity.

\section{Conclusion}
In this paper, we present a contrastive self-supervised learning method for graph transformers. We also propose a node dilation and shrinkage strategy to enhance contrastive learning. Extensive experiments with our proposed method demonstrate that it significantly improves the autism detection capability, achieving state-of-the-art performance in autism detection in ABIDE data.

\bibliographystyle{IEEEbib}
\bibliography{refs}

\end{document}